\documentclass[preprint]{sig-alternate}

\usepackage{amsmath}
\usepackage{algorithm2e}
\usepackage[breaklinks=true]{hyperref}
\usepackage{graphicx}
\usepackage{color}
\usepackage{listings}
\usepackage{listingsutf8}
\usepackage{multirow}
\usepackage{booktabs}
\usepackage{stmaryrd}

\usepackage{dcolumn}
\newcolumntype{d}[1]{D{.}{.}{#1} }

\DeclareMathOperator*{\argmin}{arg\,min}
\DeclareMathOperator*{\argmax}{arg\,max}
\DeclareMathOperator*{\sign}{sign}

\begin{document}
\conferenceinfo{KDD}{'15 Sydney, Australia}
\title{Text Segmentation based on Semantic Word Embeddings}
\numberofauthors{2} 
\author{
\alignauthor
Alexander A Alemi\\
       \affaddr{Dept of Physics\\ Cornell University}\\
       \email{aaa244@cornell.edu}
\alignauthor Paul Ginsparg\\
       \affaddr{Depts of Physics and Information Science\\ Cornell University}\\
       \email{ginsparg@cornell.edu}
}

\date{\today}

\maketitle
\begin{abstract}
    We explore the use of semantic word embeddings \cite{word2vec, glove, yoav}
in text segmentation algorithms, 
including the C99 segmentation algorithm \cite{choi1,choi2} and new
algorithms inspired by the distributed word vector representation. 
By developing a general
framework for discussing a class of  
segmentation objectives, we  study the
effectiveness of greedy versus exact optimization approaches and suggest a new
iterative refinement technique for improving the performance of greedy strategies.
We compare our results to known benchmarks
\cite{riedl,misra,choi1,choi2}, using known metrics \cite{beeferman,
windowdiff}. We demonstrate state-of-the-art performance for an untrained
method with our Content Vector Segmentation (CVS) on the Choi test set.
Finally, we apply the segmentation procedure to an in-the-wild dataset
consisting of text extracted from scholarly articles in the arXiv.org database.

\end{abstract}




\pagenumbering{arabic}

\section{Introduction}

Segmenting text into naturally coherent sections has many useful
applications in information retrieval and automated text summarization,
and has received much past attention.  An early text segmentation
algorithm was the TextTiling method introduced by Hearst \cite{hearst} in
1997.  Text was scanned linearly, with a coherence calculated for each
adjacent block, and a heuristic was used to determine the locations of
cuts.  In addition to linear approaches, there are text segmentation
algorithms that optimize some scoring objective.  An early algorithm in
this class was Choi's C99 algorithm \cite{choi1} in 2000, which also
introduced a benchmark segmentation dataset used by subsequent work.
Instead of looking only at nearest neighbor coherence, the C99 algorithm
computes a coherence score between all pairs of elements of
text,\footnote{By `elements', we mean the pieces of text combined in order
to comprise the segments. In the applications to be considered, the basic
elements will be either sentences or words.} and searches for a text
segmentation that optimizes an objective based on that scoring by greedily
making a succession of best cuts.  Later work by Choi and collaborators
\cite{choi2} used distributed representations of words rather than a bag
of words approach, with the representations generated by LSA \cite{lsa}.
In 2001, Utiyama and Ishahara introduced a statistical model for
segmentation and optimized a posterior for the segment boundaries.  Moving
beyond the greedy approaches, in 2004 Fragkou et al.\ \cite{fragkou}
attempted to find the optimal splitting for their own objective using
dynamic programming.  More recent attempts at segmentation, including
Misra et al.\ \cite{misra} and Riedl and Biemann \cite{riedl}, used LDA based
topic models to inform the segmentation task. Du et al.\ consider structured
topic models for segmentation \cite{du}. Eisenstein and Barzilay \cite{bayesseg} 
and Dadachev et al.\ \cite{automatic} both consider a Bayesian approach to text
segmentation. Most similar to our own work, Sakahara et al.\ \cite{word2vecseg} 
consider a segmentation algorithm which does affinity propagation clustering
on text representations built from word vectors learned from word2vec \cite{word2vec}.

For the most part, aside from \cite{word2vecseg}, the non-topic model based
segmentation approaches have been based on relatively simple representations of
the underlying text.  Recent approaches to learning word vectors, including
Mikolov et al.'s word2vec \cite{word2vec}, Pennington et al.'s GloVe
\cite{glove} and Levy and Goldberg's pointwise mutual information \cite{yoav},
have seen remarkable success in solving analogy tasks, machine translation
\cite{machinetrans}, and sentiment analysis \cite{glove}.  These word vector
approaches attempt to learn a log-linear model for word-word co-occurrence
statistics, such that the probability of two words $(w,w')$ appearing near one
another is proportional to the exponential of their dot product,
\begin{equation}
P(w | w') = \frac{ \exp ( w \cdot w' ) }{ \sum_v \exp ( v \cdot w' ) }\ .
\end{equation}
The method relies on these word-word co-occurrence statistics encoding  meaningful
semantic and syntactic relationships. Arora et al.\ \cite{genmodel} have shown
how the remarkable performance of these techniques can be understood in terms
of relatively mild assumptions about corpora statistics, which in turn can be
recreated with a simple generative model.

Here we explore the utility of word vectors for text segmentation,
both in the context of existing algorithms such as C99,
and when used to construct new segmentation
objectives based on a generative model for segment formation.
We will first construct a framework for describing a family of segmentation
algorithms, then discuss the specific algorithms to be investigated in detail.
We then apply our modified algorithms both to the standard Choi test set
and to a test set generated from  arXiv.org research articles.

\section{Text Segmentation}

The segmentation task is to split a text into contiguous coherent sections.
We  first build a  \emph{representation} of the text, by splitting it into $N$ basic
elements, $\vec {V_i}$ ($i=1,\ldots,N$), each a $D$-dimensional feature vector
$V_{i\alpha}$ ($\alpha=1,\ldots,D$) representing the element.
Then we assign a \emph{score} $\sigma(i,j)$ to each candidate segment, comprised of the $i^{\rm th}$
through $(j-1)^{\rm th}$ elements, and finally determine how to \emph{split} the text
into the appropriate number of segments. 

Denote a segmentation of text into $K$ segments as a list of $K$ indices $s =
(s_1, s_1, \cdots, s_{K})$, where the $k$-th segment includes the elements
$\vec V_i$ with $s_{k-1} \leq i < s_k$, with $s_{0}\equiv0$.  For example, the
string ``aaabbcccdd" considered at the character level would be properly split
with $s= (3,5,8,10)$ into (``aaa", ``bb", ``ccc", ``dd").

\subsection{Representation}

The text  representation thus amounts to turning a plain text document $T$ into an
$(N\times D)$-dimensional matrix $\mathbf{V}$, with $N$  the number of
initial elements to be grouped into coherent segments and $D$ the
dimensionality of the element representation.
For example, if segmenting at the word level then $N$ would
be the number of words in the text, and each word might be represented by a $D$-dimensional
vector, such as those obtained from GloVe \cite{glove}.
If segmenting instead  at the sentence level, then $N$ is the number
of sentences in the text and we must decide how to represent each sentence.

There are additional preprocessing decisions, for example 
using a stemming algorithm or  removing stop words before forming the representation.
Particular preprocessing decisions can have a large effect on the performance of segmentation
algorithms, but for discussing scoring functions and splitting methods
those decisions can be abstracted into the specification of the $N \times D$ matrix $\mathbf{V}$.

\subsection{Scoring}

Having built an initial representation the text, we next specify the coherence
of a segment of text with a scoring function $\sigma(i,j)$, which acts on the
representation $\mathbf{V}$ and returns a score for the segment running from $i$
(inclusive) to $j$ (non-inclusive).  The score can be a simple scalar or more general object.
In addition to the scoring function, we need to specify how to return
an aggregrate score for the entire segmentation.  This \emph{score aggregation}
function $\oplus$ can be as simple as adding the scores for the individual
segments, or again some more general function.  
The score $S(s)$ for an overall segmentation 
is given by aggregating the scores of all of the segments in the segmentation:
\begin{equation} 
    S(s) = \sigma(0, s_1) \oplus \sigma(s_1, s_2) \oplus \cdots \oplus \sigma(s_{K-1}, s_{K} )\ .
\end{equation}

Finally, to frame the segmentation
problem as a form of optimization, we  need to map the aggregated
score to a single scalar. The \emph{key} function ($\llbracket \cdot\rrbracket$)
returns this single number, so that the cost for the above segmentation is
\begin{equation}
    C(s) = \left\llbracket S(s) \right\rrbracket\ .
\end{equation}

For most of the segmentation
schemes to be considered, the score function itself
returns a scalar, so the score aggregation function $\oplus$ will be taken as simple
addition with the key function the identity, but the generality here
allows us to incorporate the C99 segmentation algorithm
\cite{choi1} into the same framework.

\subsection{Splitting}

Having specified the representation of the text and scoring of the candidate
segments, we need to prescribe how to  choose the final segmentation.  In this
work, we consider three methods: (1) greedy splitting, which at each step
inserts the best available segmentation boundary; (2) dynamic programming based
segmentation, which uses dynamic programming to find the optimal segmentation;
and (3) an iterative refinement scheme, which starts with the greedy
segmentation and then adjusts the boundaries to improve performance.

\subsubsection{Greedy Segmentation}

The greedy segmentation approach builds up a segmentation into $K$ segments by
greedily inserting new boundaries at each step to minimize the aggregate score:
\begin{align}
    s^0 &= \{ N \} \\
    s^{t+1} &= \argmin_{i \in [1,N)} C( s^t \cup \{i\} )
\end{align}
until the desired number of splits is reached.  Many published text
segmentation algorithms are greedy in nature, including the original C99
algorithm \cite{choi1}.  

\subsubsection{Dynamic Programming}

The greedy segmentation algorithm is not guaranteed to find the optimal
splitting, but dynamic programming methods can be used for
the text segmentation problem formulated in terms of optimizing a
scoring objective. For a detailed account of dynamic programming
and segmentation in general, see the thesis by Terzi \cite{terzi}. 
Dynamic programming as been applied to text segmentation in
Fragkou et al.\  \cite{fragkou}, with much success, 
but we will also consider here an optimizaton of the
the C99 segmentation algorithm using a dynamic programming approach.

The goal of the dynamic programming approach is to split the segmentation
problem into a series of smaller segmentation problems,  by expressing
the optimal segmentation of the first $n$ elements of the sequence into $k$
segments in terms of the best choice for the last segmentation boundary.
The aggregated score $S(n,k)$ for this optimal segmentation should
be minimized with respect to the key function $\llbracket \cdot \rrbracket$:
\begin{align}
    S(n,1) &= \sigma(0,n) \\
    S(n,k) &= \min_{l < n}^{\llbracket \cdot \rrbracket}  S(l,k-1) \oplus \sigma(l,n)\ .
\end{align}

While the dynamic
programming approach yeilds the optimal segmentation for our decomposable
score function, it can be costly to compute, especially for long texts.
In practice, both the optimal segmentation score and the resulting segmentation
can be found in one pass by building up a table of segmentation scores and
optimal cut indices one row at a time. 

\subsubsection{Iterative Relaxation}

Inspired by the popular Lloyd algorithm for $k$-means, we attempt to retain
the computational benefit of the greedy segmentation approach, 
but realize additional performance gains by iteratively refining the segmentation.
Since text segmentation problems require contiguous blocks of text, a natural
scheme for relaxation is to try to move each segment boundary optimally while
keeping the edges to either side of it fixed:
\begin{align}
    s^{t+1}_k &= \argmin^{\llbracket \cdot \rrbracket}_{l \in (s^t_{k-1},\ s^t_{k+1} ) } \left( \sigma(0, s^t_1) \oplus \cdots \right. \nonumber\\ 
  &\quad \left. {} \oplus \sigma(s^t_{k-1}, l) \oplus \sigma(l, s^t_{k+1}) \oplus \cdots \oplus\sigma(s^t_{K-1}, s^t_{K}) \right) \\
  &= \argmin^{\llbracket \cdot \rrbracket}_{l \in (s^t_{k-1},\ s^t_{k+1} ) } S\left( s^t - \{ s^t_k \} \cup \{ l \} \right)
\end{align}
We will see in practice that by 20 iterations it has typically converged to a fixed point 
very close to the optimal dynamic programming segmentation.

\section{Scoring Functions}

In the experiments
to follow, we will test  various choices for the representation, scoring
function, and splitting method in the above general framework.
The segmentation algorithms to be considered fall into three groups:

\subsection{C99 Segmentation}

Choi's C99 algorithm \cite{choi1} was an early text segmentation algorithm
with promising results.  The feature vector for an element of text is chosen as the pairwise cosine distances with
other elements of text, where those elements in turn are represented by a bag of stemmed words vector (after
preprocessing to remove stop words):
\begin{equation}
A_{ij} = \frac{ \sum_{w} f_{i,w} f_{j,w} }{ \sqrt{ \sum_w f_{i,w }^2 \sum_w f_{j,w}^2 } }\ ,
\label{eqn:choicosine}
\end{equation}
with $f_{i,w}$  the frequency of word $w$ in element $i$.
The pairwise cosine distance matrix is noisy for these features,
and since only the relative values are meaningful, C99 employs a ranking
transformation, replacing each value of the matrix  by the fraction
of its neighbors with smaller value:
\begin{equation}
    V_{ij} = \frac{1}{r^2-1} \sum_{\scriptstyle i-r/2 \le l \leq i+r/2\atop \scriptstyle l\neq i} 
    \ \sum_{\scriptstyle  j-r/2 \leq m \leq j+r/2\atop \scriptstyle m\neq j } \left[ A_{ij} > A_{lm}\right]\ ,
\end{equation}
where the neighborhood is an $r\times r$ block around the entry, the square brackets mean 1 if the inequality is satisfied otherwise 0
(and values off the end of the matrix are not counted in the sum, or towards the normalization).  Each element of the
text in the C99 algorithm is represented by a rank transformed vector of its cosine distances to each other element.

The score function describes the average intersentence similarity
by taking the overall score to be
\begin{equation}
C(s) = \frac{ \sum_k \beta_k }{ \sum_k \alpha_k }\ ,
\end{equation} 
where $\beta_k = \sum_{s_{k-1} \leq i < s_k} \sum_{s_{k-1} \leq j < s_k} V_{ij}$ is the sum
of all ranked cosine similarities in a segment and $\alpha_k =
(s_{k+1}-s_{k})^2 $ is the squared length of the segment. This score function is still decomposable, but
 requires that we define the local score function to return a pair,
\begin{equation}
\sigma(i,j) = \Bigl( \sum_{i \leq k < j} \sum_{i \leq k < j} V_{ij} ,\ (j-i)^2 \Bigr)\ ,
\end{equation}
with  score aggregation function defined as component addition,
\begin{equation}
(\beta_1, \alpha_1) \oplus ( \beta_2,\ \alpha_2 )  = ( \beta_1 + \beta_2,\ \alpha_1 + \alpha_2 )\ ,
\end{equation}
and  key function defined as division of the two components,
\begin{equation}
\llbracket (\beta, \alpha) \rrbracket = \frac{\beta}{\alpha}\ .
\end{equation}

While earlier work with the C99 algorithm considered only a greedy splitting approach, 
in the experiments that follow we will use our more general framework to 
explore both optimal dynamic programming and refined iterative
versions of C99.
Followup work by Choi et al.\ \cite{choi2} explored the effect
of using combinations of LSA word vectors in eq.~(\ref{eqn:choicosine}) in place of the $f_{i,w}$.  
Below we will explore the effect of using combinations of word vectors to represent the elements.

\subsection{Average word vector}

To assess the utility of word vectors in segmentation,
we  first investigate how they can be used to improve the C99 algorithm,
and then consider more general scoring functions based on our word vector
representation. As the representation of an element, we take
\begin{equation}
V_{ik} = \sum_w f_{iw} v_{wk}\ ,
\end{equation}
with $f_{iw}$ representing the frequency of word $w$ in element $i$, and $v_{wk}$ representing
the $k^{\rm th}$ component of the word vector for word $w$ as learned by a word vector
training algorithm, such as word2vec \cite{word2vec} or GloVe \cite{glove}.

The length of word vectors varies strongly across the vocabulary and in general
correlates with word frequency. In order to mitigate the effect of common words,
we will sometimes weight the sum by the inverse document frequency  (idf) of the word in the corpus:
\begin{equation}
V_{ik} = \sum_w f_{iw} \log \frac{ |D|}{df_w} v_{wk}\ ,
\end{equation}
where $df_w$ is the number of documents in which word $w$ appears. 
We can instead normalize the word vectors before adding them together
\begin{equation}
V_{ik} = \sum_w f_{iw} \tilde v_{wk} \quad \tilde v_{wk} = \frac{ v_{wk} }{ \sqrt{ \sum_k v_{wk}^2 }}\ ,
\end{equation}
or both weight by idf and normalize.

Segmentation is a form of clustering, so a natural choice for
scoring function is the sum of square deviations from the mean of the segment,
as used in $k$-means:
\begin{align}
    &\sigma(i,j) = \sum_{l}\sum_k \bigl( V_{lk} - \mu_k(i,j) \bigr)^2 \\
    &{\rm where\ }\mu_k(i,j) = \frac{1}{j-i} \sum_{l=i}^{j-1} V_{lk} \ ,  \label{eq:eucscore}
\end{align}
and which we call the Euclidean score function.
Generally, however, cosine similarity is used for word vectors,
making angles between words more important than distances.
In some experiments, we therefore normalize the word vectors first,
so that a euclidean distance score better approximates the cosine distance
(recall $\left| \tilde v - \tilde w \right|^2_2 = \left| \tilde v \right|^2_2 + \left| \tilde w \right|_2^2 - 2 \tilde v
\cdot \tilde w  = 2 ( 1 - \tilde v\cdot \tilde  w)$ for normalized vectors).

\subsection{Content Vector Segmentation (CVS)}
\label{subsec:cvs}

Trained word vectors have a remarkable amount of structure.  Analogy
tasks such as man:woman::king:? can be solved by finding the vector closest to the linear
query:
\begin{equation}
v_{\text{woman}} - v_{\text{man}} + v_{\text{king}}\ .
\end{equation}
Arora et al.\ \cite{genmodel} constructed a generative model of text
that explains how this linear structure arises and can be maintained even in
relatively low dimensional vector models.
The generative model consists of a content vector which undergoes a random
walk from a stationary distribution defined to be the product distribution on
each of its components $c_k$, uniform on the interval $[-\frac{1}{\sqrt D},
\frac{1}{\sqrt D}]$ (with $D$ the dimensionality of the word vectors).  At each point in 
time, a word vector is generated by the content vector according to a log-linear model:
\begin{equation}
P(w|c) = \frac{1}{Z_c} \exp (w \cdot c)\ , \quad Z_c = \sum_v \exp (v \cdot c)\ .
\end{equation}
The slow drift of the content vectors helps to ensure that nearby words 
obey with high probability a log-linear model for their co-occurence probability:
\begin{equation}
 \log P(w,w') = \frac{1}{2d} \left\| v_w + v_{w'} \right\|^2 - 2 \log Z \pm o(1)\ ,
\end{equation}
for some fixed $Z$. 

To segment text into coherent sections, we will boldly assume that the content vector in each 
putative segment is constant, and measure the log likelihood that all words in the segment
are drawn from the same content vector $c$. (This is similar in spirit to the
probabilistic segmentation technique proposed by Utiyama and Isahara
\cite{u00}.) Assuming the word draws $\{w_i\}$ are independent, we have that the log likelihood
\begin{equation} 
\log P(\{w_i\} | c) = \sum_i \log P(w_i|c) \propto \sum_i w_i \cdot c
\end{equation}
is proportional to the sum of the dot products of the word vectors $w_i$ with the content vector $c$.
We use a maximum likelihood estimate for the content vector:
\begin{align}
    c &= \argmax_c \log P( c| \{w_i\} ) \\
    &= \argmax_c \Bigl(\log P( \{w_i \} | c) + \log P(c) - \log P(\{w_i\} )\Bigr) \\
    &\propto \argmax_c \sum w_i \cdot c  \quad {\rm s.t.}\quad -\frac{1}{\sqrt D} < c_k < \frac{1}{\sqrt{D}}\ .
\end{align}
This determines what we will call the Content Vector Segmentation (CVS) algorithm,
based on the score function
\begin{equation} 
    \sigma(i,j) = \sum_{i \leq l < j}  \sum_k w_{lk} c_k(i,j)\ .  \label{eqn:cvsscore}
\end{equation}
The score $\sigma(i,j)$ for a segment $(i,j)$ is the sum of the dot products of the word vectors $w_{l k}$ 
with the maximum likelihood content vector $c(i,j)$ 
for the segment, with components given by
\begin{equation}
    c_k(i,j) = \sign{ \left( \sum_{i \leq l < j} w_{l,k} \right)}  \frac{1}{\sqrt D}\ .
\end{equation}
  The maximum likelihood content vector  thus has components $\pm\frac{1}{\sqrt D}$,
depending on whether the sum of the word
vector components in the segment is positive or negative.

This score function will turn out to generate some of the most
accurate segmentation results. Note that CVS is completely untrained with respect
to the specific text to be segmented, relying only
on a suitable set of word vectors, derived from some corpus in the language of choice.
While CVS is most justifiable when working on the word vectors directly, we will also explore the effect of
normalizing the word vectors before applying the objective.

\section{Experiments}

To explore the efficacy of different segmentation strategies and
algorithms, we performed segmentation experiments on two datasets. The first is
the Choi dataset \cite{choi1}, a common benchmark used in earlier segmentation
work, and the second is a similarly constructed dataset based on articles
uploaded to the arXiv, as will be described in Section \ref{subsc:arxiv}.
All code and data used for these experiments is available
online\footnote{\url{github.com/alexalemi/segmentation}}.

\subsection{Evaluation}

To evaluate the performance of our algorithms, we use two standard
metrics: the $P_k$ metric and the Window\-Diff (WD) metric.
For text segmentation, near misses should get more credit than far misses.
The $P_k$ metric \cite{beeferman}, captures the probability for a probe composed of a pair of nearby
elements  (at constant distance positions $(i,i+k)$) 
to be placed in the same segment by both reference and hypothesized segmentations.
In particular, the $P_k$ metric counts the number of disagreements on the probe elements:
\begin{align}
P_k = \frac{1}{N-k}& \sum_{i=1}^{N-k} 
\bigl[ \delta_{\text{hyp}}(i,i+k) \neq \delta_{\text{ref}}(i,i+k) \bigr]\\
k & \underset{\rm {nearest\atop integer}}{=} \frac{1}{2} \frac{ \text{\# elements} }{ \text{\# segments }}-1\ ,\nonumber
\end{align}
where
$\delta(i,j)$ is equal to 1 or 0 according to whether or not
both element $i$ and $j$ are in the same segment in  hypothesized
and reference segmentations, resp., and the argument of
the sum tests agreement of the hypothesis and reference segmentations.
($k$ is taken to be one less than the integer closest to half of the number of elements
divided by the number of segments in the reference segmentation.)
The total is then divided by the total number of probes.  This 
metric counts the number of disagreements, so lower scores indicate better agreement between the two segmentations.
Trivial strategies such as choosing only a single segmentation, 
or giving each element its own segment, or giving constant boundaries or random boundaries,
tend to produce values of around 50\% \cite{beeferman}.

The $P_k$ metric has the disadvantage that it penalizes false positives more severely than false negatives,
and can suffer when the distribution of segment sizes varies.  
Pevzner and Hearst \cite{windowdiff} introduced the WindowDiff (WD)
metric:
\begin{equation}
WD = \frac{1}{N-k} \sum_{i=1}^{N-k} 
\bigl[ b_{\text{ref}}(i,i+k) \ne b_{\text{hyp}}(i,i+k) \bigr]\ ,
\end{equation}
where $b(i,j)$ counts the number of boundaries between location
$i$ and $j$ in the text, and an error is registered if the hypothesis and
reference segmentations disagree on the number of boundaries.
In practice, the $P_k$ and WD scores are highly correlated, with $P_k$ 
more prevalent in the literature --- we will provide both for most of the experiments here.

\subsection{Choi Dataset}
\label{sec:choidataset}

The Choi dataset is used to test whether a segmentation algorithm can
distinguish natural topic boundaries.  It concatenates the first $n$ sentences
from ten different documents chosen at random from a 124 document
subset of the Brown corpus 
(the \texttt{ca**.pos} and \texttt{cj**.pos} sets) \cite{choi1}.
The number of sentences $n$ taken from each document is chosen
uniformly at random within a range specified by the subset id (i.e., as min--max \#sentences).  There are four
ranges considered: (3--5, 6--8, 9--11, 3--11), the first three of which
have 100 example documents, and the last 400 documents. 
The dataset can be obtained from an archived version of
the C99 segmentation code
release\footnote{
    \url{http://web.archive.org/web/20010422042459/http://www.cs.man.ac.uk/~choif/software/C99-1.2-release.tgz}
(We thank with Martin Riedl for pointing us to the dataset.)}.
An extract from one of the documents in the test set is shown in Fig.~\ref{fig:choiexample}.

\begin{figure}[ht]
\begin{lstlisting}
==========
Some of the features of the top portions of Figure 1 and Figure 2 were mentioned in discussing Table 1 . 
First , the Onset Profile spreads across approximately 12 years for boys and 10 years for girls . 
In contrast , 20 of the 21 lines in the Completion Profile ( excluding center 5 for boys and 4 for girls ) are bunched and extend over a much shorter period , approximately 30 months for boys and 40 months for girls . 
The Maturity Chart for each sex demonstrates clearly that Onset is a phenomenon of infancy and early childhood whereas Completion is a phenomenon of the later portion of adolescence . 
==========
The many linguistic techniques for reducing the amount of dictionary information that have been proposed all organize the dictionary 's contents around prefixes , stems , suffixes , etc . 
. 
A significant reduction in the voume of store information is thus realized , especially for a highly inflected language such as Russian . 
For English the reduction in size is less striking . 
This approach requires that : ( 1 ) each text word be separated into smaller elements to establish a correspondence between the occurrence and dictionary entries , and ( 2 ) the information retrieved from several entries in the dictionary be synthesized into a description of the particular word . 
==========
\end{lstlisting} 
    \caption{Example of two segments from the Choi dataset, taken from
     an entry in the 3--5 set. Note the appearance of a ``sentence" with the
single character ``." in the second segment on line 8. 
These short sentences can confound the benchmarks. } 
\label{fig:choiexample} 
\end{figure}

\vspace{-6pt}

\subsubsection{C99 benchmark}

We will explore the effect of changing the representation and
splitting strategy of the C99 algorithm. In order to give fair
comparisons we implemented our own version of the C99 algorithm (oC99).
The C99 performance depended sensitively  on the details of the text
preprocessing.  Details can be found in Appendix \ref{app:choi}.

\subsubsection{Effect of word vectors on C99 variant}
\label{subsec:effect}

The first experiment explores the ability of word vectors to improve the performance of the C99 algorithm.
The word vectors were learned by GloVe
\cite{glove} on a 42 billion word set of the Common Crawl corpus in 300
dimensions\footnote{Obtainable from
\url{http://www-nlp.stanford.edu/data/glove.42B.300d.txt.gz}}.
We emphasize that these word vectors were not trained on the Brown or
Choi datasets directly, and instead come from a general corpus of English.
These vectors were chosen in order to isolate any improvement due to the word vectors from
any confounding effects due to details of the training procedure. The results
are summarized in Table~\ref{tab:choivec} below.
The upper section cites results from \cite{choi2}, exploring the utility of using LSA word vectors, and 
showed an improvement of a few percent over their baseline C99 implementation.
The middle section shows results from \cite{riedl},
which augmented the C99 method by representing each element with a
histogram of topics learned from LDA.  Our results are in the lower section,
showing how word vectors improve the performance of the algorithm. 

\begin{table*}[htbp]
    \centering
    \begin{tabular}{l|r|r|r|r|r|r|r|r|}
        & \multicolumn{4}{c|}{$P_k$} & \multicolumn{4}{c|}{WD} \\
        \hline
        Algorithm           & 3--5 & 6--8 & 9--11 & 3--11& 3--5 & 6--8 & 9--11 & 3--11 \\
        \hline
        C99 \cite{choi2}& 12 & 11 & 9 & 9 & \multicolumn{4}{c|}{ } \\
        C99LSA & 9 & 10 & 7 & 5 & \multicolumn{4}{c|}{ }  \\
        \hline
        C99 \cite{riedl}   & \multicolumn{4}{c|}{11.20} & \multicolumn{4}{c|}{12.07}  \\
        C99LDA & \multicolumn{4}{c|}{4.16} & \multicolumn{4}{c|}{4.89} \\
        \hline
        \hline
        oC99 & 14.22 & {\bf 12.20} & {\bf 11.59} & 15.56 & 14.22 & {\bf 12.22} & {\bf 11.60} & 15.64 \\
        oC99tf & 12.14  & 13.17 & 14.60 & 14.91 & 12.14 & 13.34 & 15.22 & 15.22\\
        oC99tfidf & {\bf 10.27} & 12.23 & 15.87 & {\bf 14.78}     & {\bf 10.27} & 12.30 & 16.29 & {\bf 14.96} \\
        oC99k50 & 20.39 & 21.13 & 23.76 & 24.33     & 20.39  & 21.34 & 23.26 & 24.63  \\
        oC99k200 & 18.60 & 17.37 & 19.42 & 20.85     & 18.60  & 17.42 & 19.60 & 20.97  \\
        \hline
    \end{tabular}

    \caption{ Effect of using word vectors in the C99 text segmentation algorithm.
$P_k$ and WD results are shown (smaller values indicate better performance).
The top section (C99 vs.\ C99LSA) shows the few percent improvement over the C99 baseline reported in \cite{choi2} 
of using LSA to encode the words.
The middle section (C99 vs.\ C99LDA) shows the effect of modifying the C99 algorithm to work on
histograms of LDA topics in each sentence, from \cite{riedl}.
The bottom section shows the effect of using word vectors trained from GloVe \cite{glove} in our
oC99 implementation of the C99 segmentation algorithm.
The oC99tf implementation sums the word vectors in each sentence, with no rank transformation, after
removing stop words and punctuation.
oC99tfidf weights the sum by the log of the inverse document frequency of each word.
The oC99k models use the word vectors to form a topic model by doing spherical $k$-means on the word vectors.
oC99k50 uses 50 clusters and oC99k200 uses 200.}

    \label{tab:choivec}
\end{table*}

In each of these last experiments, we turned off the rank transformation, pruned the stop words and
punctuation, but did not stem the vocabulary.  Word vectors can be incorporated
in a few natural ways.  Vectors for each word in a sentence
can simply be summed, giving results shown in the oC99tf row.  
But all words are not created equal, so the sentence representation might be dominated by
the vectors for common words.
In the oC99tfidf row, the word vectors are weighted by $\text{idf}_i = \log \frac{ 500 }{ \text{df}_i }$
(i.e.,  the log of the inverse document frequency of each word in the Brown corpus, which has
500 documents in total) before summation.
We see some improvement from using word vectors, for example the $P_k$ of 14.78\% 
for the oC99tfidf method on the 3--11 set, compared to $P_k$ of 15.56\%  for our baseline C99 implementation.
On the shorter 3--5 test set, our oC99tfidf method achieves $P_k$ of 10.27\%  versus the baseline oC99 $P_k$ of 14.22\% .
To compare to the various topic model based approaches, e.g.\ \cite{riedl},
we perform spherical $k$-means clustering on the word vectors \cite{ng}
and represent each sentence as a histogram of its word clusters (i.e., as a vector in the space of clusters, with components equal
to the number of its words in that that cluster).
In this case, the word topic representations (oC99k50 and oC99k200 in Table~\ref{tab:choivec}) do not perform as well 
as the C99 variants of \cite{riedl}.
But as was noted in \cite{riedl}, those topic models 
were trained on cross-validated subsets of the Choi dataset,
and benefited from seeing virtually all of the sentences in the test sets already in each training set,
so have an unfair advantage that would not necessarily convey to real world applications.
Overall, the results in Table~\ref{tab:choivec} illustrate that the word vectors obtained from 
GloVe can markedly improve existing segmentation algorithms.

\subsubsection{Alternative Scoring frameworks}

The use of word vectors permits consideration of natural  scoring functions other than C99-style segmentation scoring.
The second experiment examines
 alternative scoring frameworks using the same GloVe word vectors as in the previous experiment.
 To test the utility of the scoring functions more directly, for these experiments we used the optimal dynamic
programming segmentation.  Results are summarized in Table~\ref{tab:choiscore},
which shows the average $P_k$ and WD scores on the 3--11
subset of the Choi dataset.  In all cases, we removed stop words and punctuation,
did not stem, but after preprocessing removed sentences with fewer than 5
words.

\begin{table}
    \centering
    \begin{tabular}{r|c|c|r|r|}
        Algorithm & rep & n & $P_k$ & WD \\
        \hline
        \multirow{2}{*}{oC99} & tf & - & 11.78 & 11.94 \\
        \cline{2-5}
         & tfidf & - & 12.19 & 12.27 \\
        \hline
        \multirow{4}{*}{Euclidean} & \multirow{2}{*}{tf} & F & 7.68 & 8.28 \\
        \cline{3--5}
         & & T & 9.18 & 10.83\\
        \cline{2-5}
         & \multirow{2}{*}{tfidf} & F & 12.89 & 14.27 \\
        \cline{3--5}
         &  & T & 8.32 & 8.95 \\
        \hline
        \multirow{4}{*}{Content (CVS)} & \multirow{2}{*}{tf} & F & 5.29 & 5.39 \\
        \cline{3--5}
         & & T & 5.42 & 5.55 \\
        \cline{2-5}
         & \multirow{2}{*}{tfidf} & F & 5.75 & 5.87 \\
        \cline{3--5}
         &  & T & 5.03 & 5.12 \\
        \hline
    \end{tabular}

    \caption{Results obtained by varying the scoring function. These
        runs were on the 3--11 set from the Choi database, with a word cut of 5
        applied, after preprocessing to remove stop words and punctuation, but
        without stemming.  The CVS method does remarkably better than
        either the C99 method or a Euclidean distance-based scoring function.
    }

    \label{tab:choiscore}
\end{table}

Note first that the dynamic programming results for our implementation of C99
with tf weights gives $P_k=11.78\%$, 3\% better than the 
greedy version result of 14.91\% reported in Table~\ref{tab:choivec}.
This demonstrates that the original C99 algorithm and its applications can
benefit from a more exact minimization than given by the greedy
approach.  We considered two natural score functions: the Euclidean scoring
function (eqn.~(\ref{eq:eucscore})) which minimizes the sum of the square deviations of each vector
in a segment from the average vector of the segment,
and the Content Vector scoring (CVS)  (eqn.~(\ref{eqn:cvsscore}) of section~\ref{subsec:cvs}),
which uses an approximate log posterior for the words in
the segment, as determined from its maximum likelihood content vector.
In each case, we consider vectors for each sentence generated
both as a strict sum of the words comprising it (tf approach), and as a sum
weighted by the log idf (tfidf approach, as in sec.~\ref{subsec:effect}).
Additionally, we consider
the effect of normalizing the element vectors before starting the score minimization,
as indicated by the $n$ column.

The CVS score function eqn.~(\ref{eqn:cvsscore}) performs the best overall, 
with $P_k$ scores below 6\%,
indicating an improved segmentation performance using a score function
adapted to the choice of representation.
While the most principled score
function would be the Content score function using tf weighted element vectors
without normalization, the normalized tfidf scheme actually performs the best.
This is probably due to the uncharacteristically large effect common words have
on the element representation, which the log idf weights and the normalization
help to mitigate.

Strictly speaking, the idf weighted schemes cannot claim to be completely
untrained, as they benefit from word usage statistics in the Choi test set, but
the raw CVS method still demonstrates a marked improvement on the 3--11 subset,
5.29\% $P_k$ versus the optimal C99 baseline of 11.78\% $P_k$.

\subsubsection{Effect of Splitting Strategy}

To explore the effect of the splitting strategy and to compare with our overall
results on the Choi test set against other published benchmarks, in our third experiment 
we ran the raw CVS method
against all of the Choi test subsets,
 using all three splitting strategies discussed: greedy, refined, and
dynamic programming.  These results are summarized in Table~\ref{tab:choibest}.

\begin{table}
    \centering
    \begin{tabular}{l|d{2}|d{2}|d{2}|d{2}|}
        Alg & \multicolumn{1}{c|}{3--5} & \multicolumn{1}{c|}{6--8} & \multicolumn{1}{c|}{9--11} & \multicolumn{1}{c|}{3--11} \\
        \hline
        TT \cite{choi1} & 44 & 43 & 48 & 46 \\
        C99 \cite{choi1} & 12 & 9 & 9 & 12 \\
        C01 \cite{choi2} & 10 & 7 & 5 & 9 \\
        U00 \cite{u00} & 9 & 7 & 5 & 10 \\
        F04 \cite{fragkou}& 5.5 & 3.0 & 1.3 & 7.0 \\
        \hline
        G-CVS  & 5.14 & 4.82 & 6.38 & 6.49 \\
        R-CVS  & 3.92 & 3.75 & 5.17 & 5.65 \\
        DP-CVS & 3.41 & 3.45 & 4.45 & {\bf 5}.\protect\llap{\bf.}{\bf 29} \\
        \hline
        M09 \cite{misra}& 2.2 & 2.3 & 4.1 & 2.3 \\
        R12 \cite{riedl}& 1.24 & 0.76 & 0.56 & 0.95 \\
        D13 \cite{du}& 1.0 & 0.9 & 1.2 & 0.6 \\
         \hline
 \end{tabular}

    \caption{Some published $P_k$ results on the Choi dataset against our 
raw CVS method.  G-CVS uses a greedy splitting strategy, R-CVS uses up to 20 iterations to refine the 
results of the greedy strategy, and DP-CVS shows the optimal results obtained by dynamic programming.
We include the topic modeling results M09, R12, and D13 for reference, but for reasons detailed in the text do not
regard them as comparable, due to their mingling of test and training samples.
}

    \label{tab:choibest}
\end{table}

Overall, our method outperforms all previous untrained methods.
As commented regarding Table~\ref{tab:choivec} (toward the end of subsection~\ref{subsec:effect}), we have included the results of the 
topic modeling based approaches M09 \cite{misra}, R12 \cite{riedl}, and D13 \cite{automatic} for reference.
But due to repeat appearance of the same sentences throughout each section of the Choi dataset, methods that split that dataset into test and training sets have unavoidable
access to the entirety of the test set during training, albeit in different order.\footnote{In \cite{riedl}, it is observed that
``This makes the Choi data set artificially easy for supervised approaches.'' See appendix~\ref{app:choit}.}
These results can therefore only  be compared to other algorithms permitted to make extensive use of the test data during cross-validation training.
Only the TT, C99, U00 and raw CVS method can be considered as completely untrained.
The C01 method derives its LSA vectors from the Brown corpus, from which the Choi test set is constructed, but that provides only a weak benefit,
and the F04 method is additionally trained on a subset of the test  set to achieve its best performance, but its use only of idf values provides a similarly weak benefit.

\begin{figure}
    \centering
    \includegraphics[width=1\linewidth]{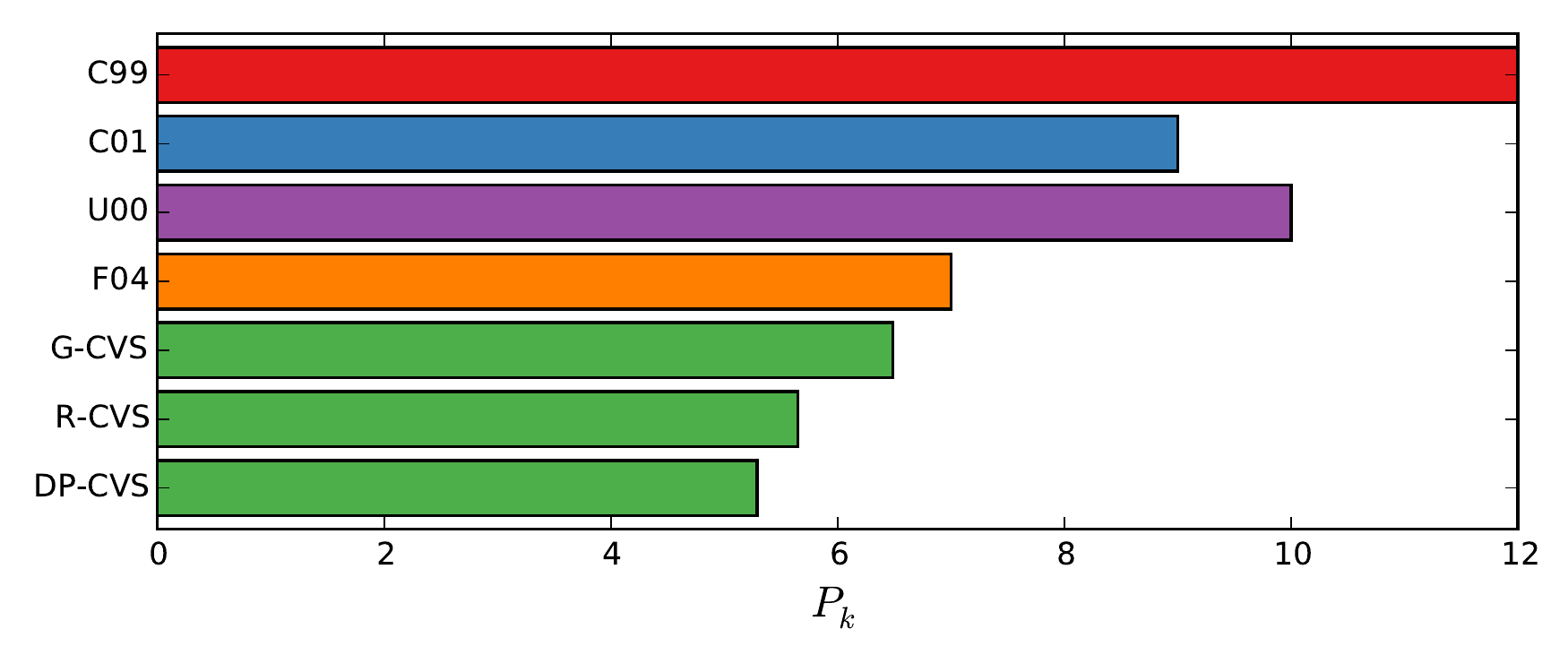}
    \caption{Results from last column of Table~\ref{tab:choibest} reproduced to highlight the performance of the CVS segmentation algorithm compared to similar untrained algorithms.
    Its superior performance in an unsupervised setting suggests applications on documents ``in the wild''.}
\label{fig:results}
\end{figure}

We emphasize that the raw CVS method is completely independent of the Choi test set, using word vectors derived from a completely different corpus.
In Fig.~\ref{fig:results}, we reproduce the relevant results from the last column of Table~\ref{tab:choivec} to highlight the performance benefits provided by the semantic word embedding.

Note also the surprising performance of the refined splitting strategy, 
with the R-CVS results in Table~\ref{tab:choibest}
much lower than the greedy G-CVS results, and moving close to the 
optimal DP-CVS results, at far lower computational cost.
In particular, taking the dynamic programming segmentation as the
true segmentation, we can assess the performance of the refined strategy.
As seen in Table~\ref{tab:choirvdp}, the refined segmentation
very closely approximates the  optimal segmentation.

This is important in practice since
the dynamic programming segmentation is much slower, taking five
times longer to compute on the 3--11 subset of the Choi test set.
The dynamic programming segmentation becomes computationally
infeasible to do at the scale of  word level segmentation on
the arXiv dataset considered in the next section,
whereas the refined segmentation method remains eminently feasible.

\begin{table}
    \centering
    \begin{tabular}{l|d{2}|d{2}|d{2}|d{2}|}
         & \multicolumn{1}{c|}{3--5} & \multicolumn{1}{c|}{6--8} & \multicolumn{1}{c|}{9--11} & \multicolumn{1}{c|}{3--11} \\
        \hline
        R-CVS vs DP-CVS \cite{choi1} & 0.90 & 0.65 & 1.16 & 1.15 \\
        \hline
    \end{tabular}

    \caption{ Treating the dynamic programming splits as the true answer, the error of the refined splits as measured in $P_k$ across the subsets of the Choi test set.}

    \label{tab:choirvdp}
\end{table}

\subsection{ArXiv Dataset}
\label{subsc:arxiv}

Performance evaluation on the Choi test set implements segmentation
at the sentence level, i.e., with segments of composed of sentences as the basic elements.
But text sources do not necessarily have well-marked sentence boundaries.  
The arXiv is a repository of scientific articles which for practical reasons extracts text
from PDF documents (typically using  \texttt{pdfminer/pdf2txt.py}). That Postscript-based format was
originally intended only as a means of formatting text on a page,
rather than as a network transmission format encoding syntactic or semantic information.
The result is often somewhat corrupted, either due to the handling of mathematical notation, the
presence of footers and headers, or even just font encoding issues.

To test the segmentation algorithms in a realistic setting, we created a test
set similar to the Choi test set, but based on text extracted from PDFs retrieved from
the arXiv database.  Each test document is composed of a random number of contiguous
words, uniformly chosen between 100 and 300, sampled at random from the text
obtained from arXiv articles.
The text was preprocessed by lowercasing and inserting spaces around every non-alphanumeric character,
then splitting on whitespace to tokenize. An example of two of the segments of the
first test document is shown in Figure \ref{fig:arxexample} below.

\begin{figure}[ht]
\begin{lstlisting}
. nature_414 : 441 - 443 . 12 seinen , i . and schram a . 2006 . social_status and group norms : indirect_reciprocity in a helping experiment . european_economic_review 50 : 581 - 602 . silva , e . r . , jaffe , k . 2002 . expanded food choice as a possible factor in the evolution of eusociality in vespidae sociobiology 39 : 25 - 36 . smith , j . , van dyken , j . d . , zeejune , p . c . 2010 . a generalization of hamilton ' s rule for the evolution of microbial cooperation science_328 , 1700 - 1703 . zhang , j . , wang , j . , sun , s . , wang , l . , wang , z . , xia , c . 2012 . effect of growing size of interaction neighbors on the evolution of cooperation in spatial snowdrift_game . chinese_science bulletin 57 : 724 - 728 . zimmerman , m . , egu `i luz , v . , san_miguel ,
of ) e , equipped_with the topology of weak_convergence . we will state some results about random measures . 10 definition a . 1 ( first two moment measures ) . for a random_variable z , taking values in p ( e ) , and k = 1 , 2 , . . . , there is a uniquely_determined measure $\mu$ ( k ) on b ( ek ) such that e [ z ( a1 ) $\cdot$_$\cdot$_$\cdot$ z ( ak ) ] = $\mu$ ( k ) ( a1 $\times$ $\cdot$_$\cdot$_$\cdot$ $\times$ ak ) for a1 , . . . , ak $\in$ b ( e ) . this is called the kth_moment measure . equivalently , $\mu$ ( k ) is the unique measure such that e [ hz , $\phi$ 1i $\cdot$_$\cdot$_$\cdot$ hz , $\phi$ ki ] = h $\mu$ ( k ) , $\phi$ 1 $\cdot$_$\cdot$_$\cdot$ $\phi$ ki , where h . , . i denotes integration . lemma a . 2 ( characterisation of deterministic random measures ) . let z be a random_variable_taking values in p ( e ) with the first two moment measures $\mu$ : = $\mu$ ( 1 ) and $\mu$ ( 2 ) . then the following_assertions_are_equivalent : 1 . there is $\nu$ $\in$ p ( e ) with z = $\nu$ , almost_surely . 2 . the second_moment measure has product - form , i . e . $\mu$ ( 2 ) = $\mu$ $\otimes$ $\mu$ ( which is equivalent to e [ hz , $\phi$ 1i $\cdot$ hz , $\phi$ 2i ] = h $\mu$ , $\phi$ 1i $\cdot$ h $\mu$ , $\phi$ 2i ( this is in fact equivalent to e [ hz , $\phi$ i2 ]
\end{lstlisting} 
    \caption{Example of two of the segments from a document in the arXiv test set. }
    \label{fig:arxexample} 
\end{figure}

This is a much more difficult segmentation task:  due to the presence of
numbers and many periods in references, there are no clear sentence boundaries
on which to initially group the text, and no natural boundaries are suggested
in the test set examples.  Here segmentation algorithms must work directly at
the ``word" level, where word can mean a punctuation mark. The presence of 
garbled mathematical formulae adds to the difficulty of making
sense of certain streams of text.

In Table \ref{tab:arx}, we summarize the results of three word vector powered
approaches, comparing a C99 style algorithm to our content vector based
methods, both for unnormalized and normalized word vectors.
Since much of the language of the scientific articles is specialized,
the word vectors used in this case were obtained from GloVe trained on
a corpus of similarly preprocessed texts from 98,392 arXiv articles.
(Since the elements are now words rather than sentences,
the only issue involves whether or not those word vectors are normalized.)
As mentioned, the dynamic programming approach is prohibitively expensive
for this dataset.

\begin{table}
    \centering
    \begin{tabular}{r|l|d{2}|d{2}|}
        Alg & S &  \multicolumn{1}{c|}{$P_k$} & \multicolumn{1}{c|}{WD} \\
        \hline
        oC99 & G & 47.26 & 47.26 \\
        oC99 & R & 47.06 & 49.16 \\
        \hline
        CVS & G & 26.07 & 28.23 \\
        CVS & R & 25.55 & 27.73 \\
        \hline
        CVSn & G & 24.63 & 26.69 \\
        CVSn & R & {\bf 24}.\protect\llap{\bf .}{\bf 03} & {\bf 26}.\protect\llap{\bf .}{\bf 15} \\
        \hline
    \end{tabular}

    \caption{ Results on the arXiv test set for the C99 method using word vectors (oC99),
    our CVS method, and CVS method with normalized word vectors (CVSn).
    The $P_k$ and WD metrics are given for both the greedy (G) and refined splitting strategies (R),
    with respect to the reference segmentation in the test set. The refined strategy was
     allowed up to 20 iterations to converge. The refinement converged for
     all of the CVS runs, but failed to converge for some documents in the
    test set under the C99 method. Refinement improved performance in all cases, and our
CVS  methods improve significantly over the C99 method for this task. 
}

    \label{tab:arx}
\end{table}

We see that the CVS method performs far better on the test set than the C99
style segmentation using word vectors.
The $P_k$ and WD values obtained are not as impressive as those obtained on the Choi test set,
but this test set offers a much more challenging segmentation task: it requires
the methods to work at the level of words, and as well includes the possibility that natural
topic boundaries occur in the test set segments themselves.
The segmentations obtained with the CVS method typically appear  sensibly split
on section boundaries, references and similar formatting boundaries,
not known in advance to the algorithm.

\begin{figure} 
    \centering
    \includegraphics[width=1\linewidth]{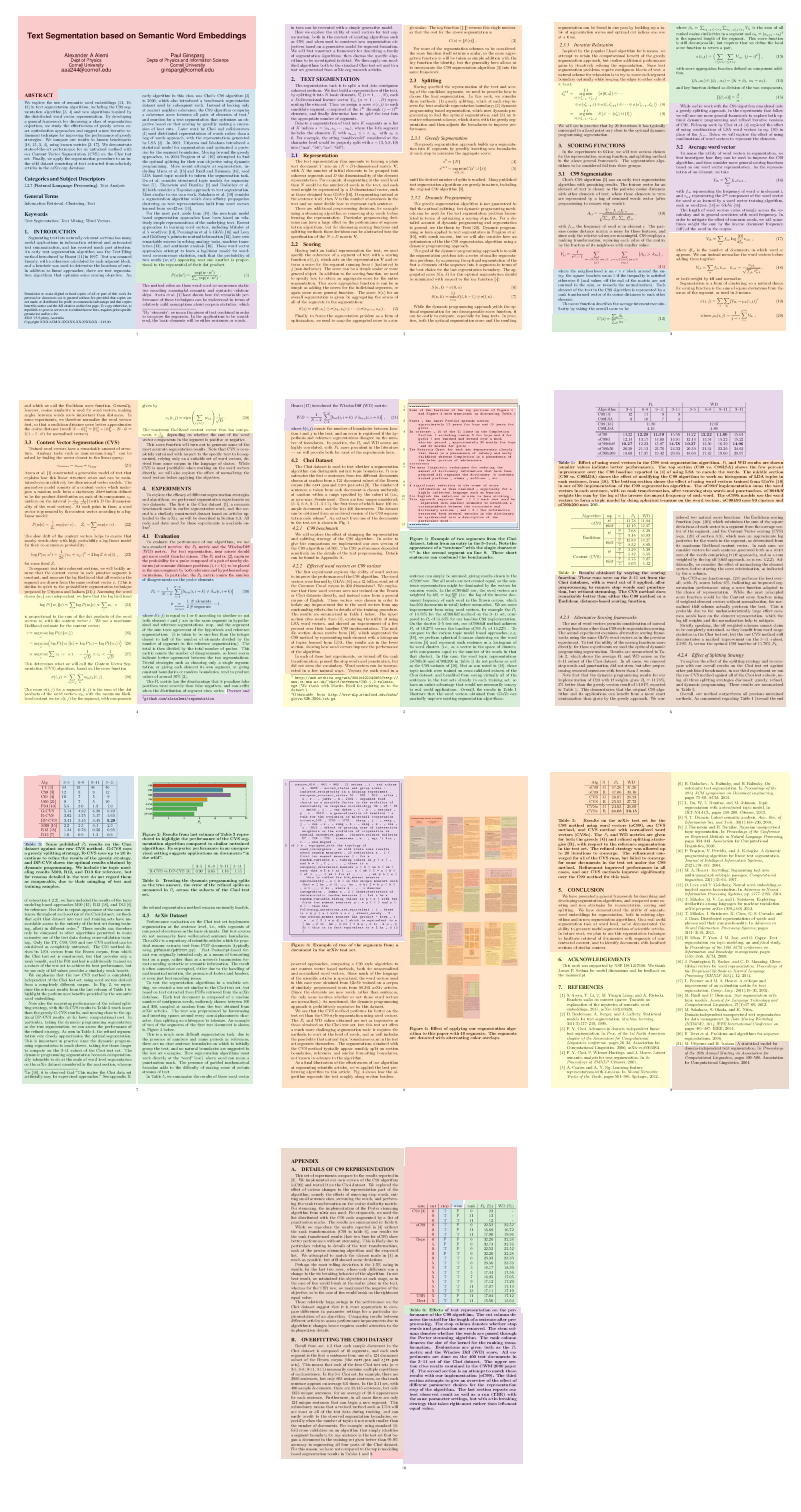}
    \caption{Effect of applying our segmentation algorithm to this paper with 40 segments.  The segments 
    are denoted with alternating color overlays. }
\label{fig:papersplit}
\end{figure}

As a final illustration of the effectiveness of our algorithm at segmenting
scientific articles, we've applied the best performing algorithm to this article.
Fig.~\ref{fig:papersplit} shows how the algorithm segments the text roughly along section borders.

\section{Conclusion}

We have presented a general framework for describing and developing 
segmentation algorithms, and compared some existing and new
strategies for representation, scoring and splitting. 
We have demonstrated the utility of semantic word embeddings for segmentation, 
both in existing algorithms and  in new segmentation algorithms.
On a real world segmentation task at word level, we've demonstrated the ability
to generate useful segmentations of scientific articles.
In future work, we plan to use this segmentation technique to facilitate retrieval of documents
with segments of concentrated content, and to identify documents with localized sections of
similar content.

\section{Acknowledgements}

This work was supported by NSF IIS-1247696. We thank James P. Sethna for useful discussions and
for feedback on the manuscript.

\bibliographystyle{habbrv}
\bibliography{segmentref} 

\newpage

\appendix
\section{Details of C99 Representation}
\label{app:choi}
This set of experiments compare to the results
reported in  \cite{choi1}.  We implemented our own
version of the C99 algorithm (oC99) and tested it on the Choi dataset.  
We explored the effect of various changes to the representation part of the
algorithm, namely the effects of removing stop words, cutting small sentence
sizes, stemming the words, and performing the rank transformation on the cosine
similarity matrix.  For stemming, the implementation of the Porter stemming
algorithm from \texttt{nltk} was used.  For stopwords, we used the list
distributed with the C99 code augmented by a list of punctuation marks. The
results are summarized in Table \ref{tab:choirep}.

While we reproduce the results reported in 
\cite{choi2} without the rank transformation (C99 in table~\ref{tab:choirep}),
our results for the rank transformed results (last two lines for oC99)
show better performance without stemming.
This is likely due to particulars relating to details of the text transformations,
such at the precise stemming algorithm  and the stopword list.
We attempted to match the choices made in
\cite{choi2}  as much as possible, but still showed some deviations.

Perhaps the most telling deviation is the 1.5\% swing in results for the last
two rows, whose only difference was a change in the tie breaking
behavior of the algorithm.  In our best result, we  minimized the objective
at each stage, so in the case of ties would break at the earlier place in the
text, whereas for the TBR row, we  maximized the negative of the
objective, so in the case of ties would break on the rightmost equal value.

These relatively large swings in the performance on the Choi dataset suggest
 that it is most appropriate to compare differences in parameter settings
for a particular implementation of an algorithm. Comparing results
between different articles to assess performance improvements due to algorithmic
changes hence requires careful attention to the implemention details.

\begin{table}[htbp]
    \centering
    \begin{tabular}{r|c|c|c|c|r|r}
        note & cut & stop & stem & rank & $P_k$ (\%) & WD (\%) \\
        \hline
        C99 \cite{choi2} & 0 & T & F & 0 & 23 & - \\
                         & 0 & T & F & 11 & 13 & - \\
                         & 0 & T & T & 11 & 12 & - \\
        \hline
        oC99 & 0 & T & F & 0 & 22.52 & 22.52 \\
                & 0 & T & F & 11& 16.69 & 16.72 \\
                & 0 & T & T & 11& 17.90 & 19.96 \\
        \hline
        
        Reps    & 0 & F & F & 0 & 32.26 & 32.28 \\
            & 5 & F & F & 0 & 32.73 & 32.76  \\
            & 0 & T & F & 0 & 22.52 & 22.52 \\
            & 0 & F & T & 0 & 32.26 & 32.28 \\
            & 0 & T & T & 0 & 23.33 & 23.33 \\
            & 5 & T & T & 0 & 23.56 & 23.59 \\
            & 5 & T & T & 3 & 18.17 & 18.30 \\
            & 5 & T & T & 5 & 17.44 & 17.56 \\
            & 5 & T & T & 7 & 16.95 & 17.05  \\
            & 5 & T & T & 9 & 17.12 & 17.20 \\
            & 5 & T & T & 11& 17.07 & 17.14  \\
            & 5 & T & T & 13& 17.11 & 17.19  \\
        \hline
         TBR   & 5 & T & F & 11& 17.04 & 17.12 \\
         Best   & 5 & T & F & 11& 15.56 & 15.64 \\
        \hline
    \end{tabular}

    \caption{Effects of text representation on the performance of the
        C99 algorithm. The cut column denotes the cutoff for the
        length of a sentence after  preprocessing.  The stop column denotes
        whether stop words and punctuation are removed.  The stem column
        denotes whether the words are passed through the Porter stemming
        algorithm. The rank column denotes the size of the kernel for the
        ranking transformation.  Evaluations are given both as the $P_k$ metric
        and the Window Diff (WD) score. All
        experiments are done on the 400 test documents in the 3--11 set of the
        Choi dataset.  The  upper section cites results contained in the CWM 2000
        paper \cite{choi2}.  The second section is an attempt to match these
        results with our implementation (oC99).  The third section attempts to give an
        overview of the effect of different parameter choices for the
        representation step of the algorithm.  The last section reports our best
    observed result as well as a run (TBR) with the same parameter settings,  but with a
    tie-breaking strategy that takes right-most rather then left-most equal value.}
    \label{tab:choirep}
\end{table}

\section{Overfitting the Choi Dataset}
\label{app:choit}

Recall from sec.~\ref{sec:choidataset} that each sample document in the Choi
dataset is composed of 10 segments, and each such segment is the first $n$
sentences from one of a 124 document subset of the Brown corpus (the
\texttt{ca**.pos} and \texttt{cj**.pos} sets).  This means that each of the
four Choi test sets ($n=$ 3-5, 6-8, 9-11, 3-11) necessarily contains multiple
repetitions of each sentence.  In the 3-5 Choi set, for example, there are 3986
sentences, but only 608 unique sentences, so that each sentence appears on
average 6.6 times.  In the 3-11 set, with 400 sample documents,  there are
28,145 sentences, but only 1353 unique sentences, for an average of 20.8
appearances for each sentence. Furthermore, in all cases there are only 124
unique sentences that can begin a new segment.  This redundancy means that a
trained method such as LDA will see most or all of the test data during
training, and can easily overfit to the observed segmentation boundaries,
especially when the number of topics is not much smaller than the number of
documents. 
For example, using standard 10-fold cross validation on an algorithm that
simply identifies a segment boundary for any sentence in the
test set that began a document in the training set gives
better than 99.9\% accuracy in segmenting all four parts of the Choi
dataset.
For this reason, we have not compared to the topic-modeling based
segmentation results in Tables \ref{tab:choivec} and \ref{tab:choibest}.

\end{document}